\pdfoutput=1
\documentclass[11pt]{article}
\usepackage{acl}
\usepackage{times}
\usepackage{latexsym}
\usepackage[T1]{fontenc}
\usepackage[utf8]{inputenc}
\usepackage{microtype}
\usepackage{amsmath}
\usepackage{braket}
\usepackage{subfigure}
\usepackage[export]{adjustbox}
\usepackage{url}
\usepackage{bm}
\usepackage{multirow}
\usepackage{tikz}
\usepackage[ruled,vlined,algo2e]{algorithm2e}
\usepackage{algorithm}
\usepackage{algorithmic}
\usepackage{graphicx}
\usepackage{setspace}
\usepackage{hyperref}
\usepackage{enumitem}
\usepackage{makecell}
\usepackage{float}

\let\Algorithm\algorithm
\renewcommand\algorithm[1][]{\Algorithm[#1]\setstretch{1}}

\usepackage{xcolor}
\usepackage[ruled,vlined,algo2e]{algorithm2e}

\SetCommentSty{mycommfont}
\AtBeginDocument{
  \let\mathbb\relax
  \DeclareMathAlphabet{\mathbb}{U}{msb}{m}{n}
}

\usepackage{amsmath}
\usepackage{booktabs}
\usepackage{amssymb}% http://ctan.org/pkg/amssymb
\usepackage{pifont}% http://ctan.org/pkg/pifont

\definecolor{myorange}{RGB}{251,229,214}
\definecolor{mygreen}{RGB}{226,240,217}

\NewDocumentCommand\cherrytitle{}{
    \includegraphics[scale=0.4]{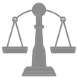}
}

\usepackage[utf8]{inputenc}

\usepackage{listings}
\definecolor{codegreen}{rgb}{0,0.5,0}
\definecolor{codeblue}{rgb}{0.25,0.5,0.5}
\definecolor{codegray}{rgb}{0.6,0.6,0.6}

\definecolor{myblue}{HTML}{4E84C4}
\definecolor{myred}{HTML}{B02418}
\definecolor{paperblue}{HTML}{077dea}
\definecolor{babyblue}{HTML}{E3EDF7} 

\newcommand{\coloredalpha}{\textcolor{paperblue}{\alpha}}
\newcommand{\coloredgamma}{\textcolor{paperblue}{\gamma}}
\newcommand{\coloredsigma}{\textcolor{paperblue}{\sigma}}
\newcommand{\coloreddelta}{\textcolor{paperblue}{\delta}}
\newcommand{\coloredmu}{\textcolor{paperblue}{\mu}}
\newcommand{\coloredbeta}{\textcolor{paperblue}{\beta}}

\makeatletter
\patchcmd{\maketitle}
 {\def\@makefnmark}
 {\def\@makefnmark{}\def\useless@macro}
 {}{}
\makeatother

\title{\textit{\texttt{Self-DC}:} When to Reason and When to Act \cherrytitle \\ Self Divide-and-Conquer for Compositional Questions}

\author{Hongru Wang$^{\coloredalpha\coloredbeta\dagger}$\thanks{$^\dagger$ Equal Contributions}, Boyang Xue$^{\coloredalpha\coloredbeta\dagger}$, Baohang Zhou$^{\coloredgamma}$, Tianhua Zhang$^{\coloredalpha}$, \\ \bf Cunxiang Wang$^{\coloredsigma}$, Huimin Wang$^{\coloredmu}$, Guanhua Chen$^{\coloreddelta\ddagger}$\thanks{$^\ddagger$ Co-corresponding Authors}, Kam-Fai Wong$^{\coloredalpha\coloredbeta\ddagger}$ \\
  $^{\coloredalpha}$The Chinese University of Hong Kong
  $^{\coloredgamma}$Nankai University \\
  $^{\coloredbeta}$MoE Key Lab of High Confidence Software Technologies, CUHK \\
  $^{\coloredsigma}$Westlake University
  $^{\coloredmu}$Jarvis Research Center, Tencent YouTu Lab \\
  $^{\coloreddelta}$Southern University of Science and Technology \\
  {\tt \{hrwang, kfwong\}@se.cuhk.edu.hk chengh3@sustech.edu.cn}
}

\begin{document}
\maketitle
\begin{abstract}
Previous research has typically concentrated on leveraging the internal knowledge of Large Language Models (LLMs) to answer known questions (i.e., \textit{internal reasoning such as generate-then-read}). In contrast, for questions that fall outside their known scope, these models rely on external knowledge retrieval to provide accurate responses (i.e., \textit{external acting such as retrieve-then-read}). However, few previous works consider the \textit{compositional questions}, which consist of several known and unknown sub-questions, necessitating the dynamic combination of previous two methods (i.e., \textit{internal reasoning and external acting}) to achieve a better trade-off between effectiveness and efficiency. To this end, we introduce a \textbf{Self} \textbf{D}ivide-and-\textbf{C}onquer (\textit{\texttt{Self-DC}}) framework, accompanying with the first \textbf{C}ompositional \textbf{u}nknown \textbf{Q}uestion-\textbf{A}nswering dataset (CuQA). This framework enables LLMs to adaptively choose between using internal knowledge and retrieving external knowledge as needed, resulting in a better trade-off between effectiveness and efficiency. Experimental results on two datasets demonstrate that \textit{\texttt{Self-DC}} can achieve comparable or even better performance with much fewer external calls compared with several strong baselines.
\end{abstract}

\section{Introduction}
Large Language Models (LLMs) \citep{ouyang2022training, touvron2023llama} possess extensive world knowledge thanks to the scaling of size of pre-training data and model \citep{kaplan2020scaling}, resulting in exceptional capabilities to answer open-domain questions using internal known knowledge encoded in their parameters \citep{genread, bang2023multitask}. However, due to the cutoff date of training data, it is difficult for them to answer questions out of their known knowledge (a.k.a., unknown questions), which necessitates the augmentation of external retrieval \citep{lewis2021retrievalaugmented, zhuang2023toolqa, vu2023freshllms, gabburo2024measuring}, such as Google Search and Wikipedia. 

\begin{figure}
    \centering
    \includegraphics[trim={5cm, 4cm, 9cm, 2cm}, clip, width=0.49\textwidth]{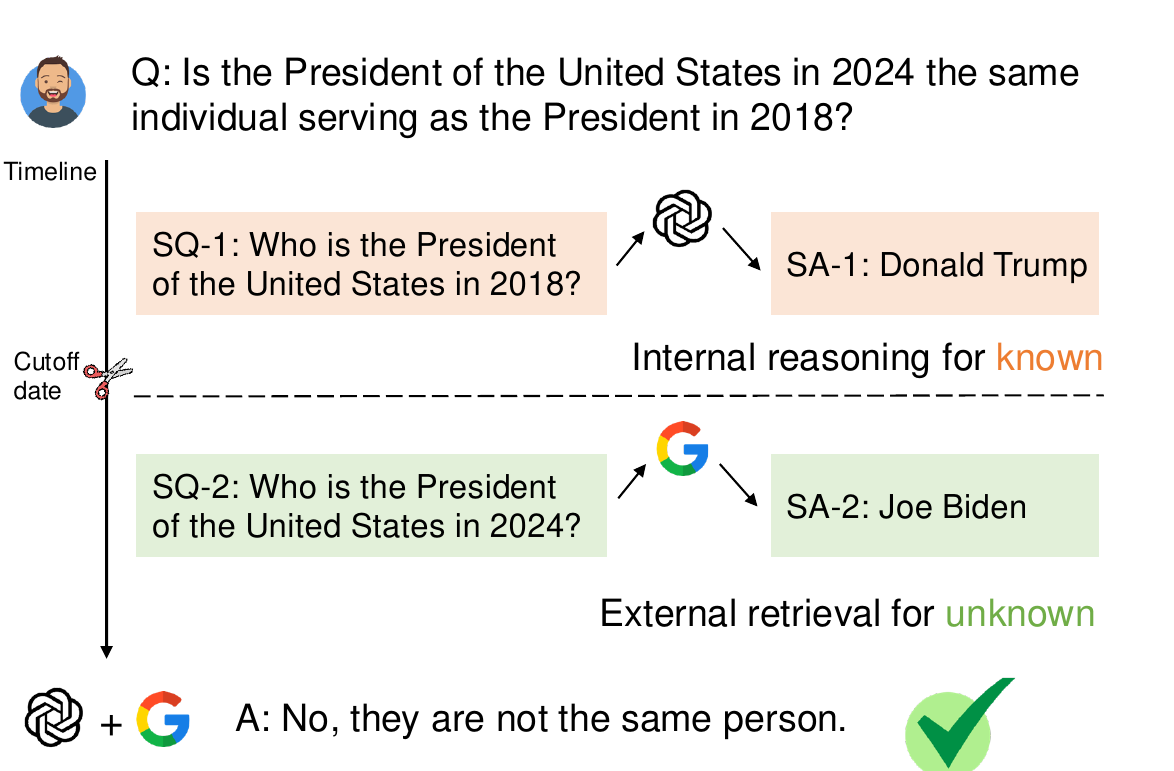}
    \caption{A example of compositional questions, in which a unknown question consists of some sub-questions can be answered using \colorbox{myorange}{\textit{known}} knowledge while other sub-questions necessitate \colorbox{mygreen}{\textit{unknown}} knowledge according to the cutoff date of LLMs.}
    \label{fig:intro_example}
    % \vspace{-6mm}
\end{figure}

To provide more accurate answers for the questions, most previous works tend to employ external retrieval methods indiscriminately without considering different types of questions, resulting in redundant retrieval and unnecessary cost \citep{trivedi-etal-2023-interleaving, shao-etal-2023-enhancing}. Alternatively, some methods simply classify questions into binary categories (i.e., known and unknown), and utilize either self-generated context or retrieved external context to answer them, respectively \citep{wang-etal-2023-self-knowledge}, following a generate-then-read \cite{genread} or retrieve-then-read \cite{lewis2021retrievalaugmented} paradigm. However, this binary classification is sub-optimal and inefficient for handling \textit{compositional questions}, which consist of multiple sub-questions where each sub-question could be known or unknown, as illustrated in Figure~\ref{fig:intro_example}. Consequently, these binary-classification methods degrade into simply retrieving information for every question, as any compositional questions containing an unknown sub-question remain entirely unknown by large language models (LLMs). Moreover, using the original compositional question as a query frequently leads to the retrieval of noisy or unrelated documents, which hinders accurate answers \cite{ma-etal-2023-query}. These limitations highlights the need for more nuanced and efficient retrieval strategies tailored to the complexity of \textit{compositional questions}.

% Both of these methods have their own limitations. On the one hand, simply calling retrieval for every question does not guarantee better performance while making the extensive known knowledge meaningless \citep{wang-etal-2023-self-knowledge}, and also brings serious efficiency issue especially when there are iterative retrieval \citep{trivedi-etal-2023-interleaving}. On the other hand, dividing the questions into binary oppositions of known and unknown oversimplifies their complexity, being infeasible or sub-optimal in certain cases. For example, as shown in Figure~\ref{fig:intro_example}, a compositional unknown question consists of both known and unknown sub-questions. Therefore, it will degrade into simply calling retrieval for every question automatically, as the compositional unknown remains an unknown question for LLMs.

In this paper, we first formally introduce \textit{compositional questions} from the perspective of known/unknown, which is more practical and challenging. 
%Specifically, 
To further specify the \textit{compositional questions}, we categorized questions into four types according to the knowledge boundaries of LLMs \footnote{The definition begins from the data side instead of model side such as the cutoff date of training data, we discuss hallucination issue of model side at Sec~\ref{sec:limitations}.}: 

% [leftmargin=*,topsep=1pt,itemsep=1pt]
\begin{itemize}
    \item \textbf{\textit{Single Known.}} The question contains no sub-questions and can be solved using internal knowledge of LLMs, such as with the generate-then-read method.
    
    \item \textbf{\textit{Single Unknown.}} The question contains no sub-questions and can only be solved using external knowledge, such as with the retrieve-then-read method.
    
    \item \textbf{\textit{Compositional Known.}} The question contains several sub-questions, and each sub-question is \textit{Single Known.}
    
    \item \textbf{\textit{Compositional Unknown.}} The question contains several sub-questions, and at least one sub-question is \textit{Single Unknown.}
\end{itemize}

% Based on this, we propose a \textbf{S}elf \textbf{D}ivide-and-\textbf{C}onquer (\textit{\texttt{Self-DC}}) approach. Recognizing the potential of LLMs in expressing their certainty/uncertainty \citep{lin2022teaching, xiao-etal-2022-uncertainty}, we divide the \textit{compositional questions} into different types of sub-questions.  We then dynamically call different functions based on the confidence score produced by the LLMs themselves. \textcolor{red}{Call function for what?} Overall, our contributions can be summarized as follows:

% build the first \textbf{C}ompositional \textbf{u}nknown \textbf{Q}uestion-\textbf{A}nswering dataset (CuQA), accompany with a automatic data collection pipeline with minimal human efforts.

Determining whether a question is known or unknown to LLMs, and whether it is a compositional question, is a complex task that may require multi-step reasoning. In this paper, we introduce a \textbf{Self} \textbf{D}ivide-and-\textbf{C}onquer (\texttt{Self-DC}), designed to effectively and efficiently identify and decompose compositional questions. The main idea of \texttt{Self-DC} is to use the inherent signals of LLM to control its own behavior, e.g., elicit the internal knowledge or call external retrieval. Specifically, we define each action as a function, and model the whole decomposition as dynamic function calls guided by self-aware confidence signals. Therefore, the internal reasoning capabilities of LLMs can be well elicited while making every external retrieval call count. In summary, our contributions can be outlined as follows:

\begin{itemize}
    \item To the best of our knowledge, we are the first to study \textit{compositional questions} from the perspective of known / unknown.
    
    \item We introduce an automatic data collection pipeline to create the first \textbf{C}ompositional \textbf{u}nknown \textbf{Q}uestion \textbf{A}nswering dataset (CuQA), serving as an important evaluation benchmark for LLMs in known/unknown.
    
    \item We present a flexible and robust \textit{\texttt{Self-DC}} framework, which is capable of adaptively calling different functions on-demand for \textit{compositional questions} decomposition.

    \item Experimental results on CuQA and FreshQA \cite{vu2023freshllms} datasets show the superiority of \texttt{Self-DC} in terms of both effectiveness and efficiency, revealing its promising potential to solve compositional reasoning problem.

\end{itemize}

\section{Related Work}

\paragraph{Known and Unknown of LLMs.}
Investigations into the known and unknown boundaries of large language models (LLMs) have gained attention in recent literature \citep{kadavath2022language, amayuelas2023knowledge, yin-etal-2023-large}. Despite the parameters of LLMs containing a wealth of knowledge to excel in various tasks, they are still limited due to the continuously increasing information. Specifically, LLMs have showcased satisfactory performance to evaluate the validity of their own claims and predict which questions they will be able to answer correctly by predicting ``P(IK)'', the probability that ``I know'' the answer to a question \citep{kadavath2022language}. Furthermore, \citet{yin-etal-2023-large} evaluate LLMs' self-knowledge by assessing their ability to identify unanswerable or unknowable questions. Similarly, \citet{amayuelas2023knowledge} further assesses the LLMs' ability to differentiate between known and unknown questions and classify them accordingly by collecting Known-Unknown Questions (KUQ). Their results show that the LLMs still have room for improvement in classifying known-vs-unknown questions, even with the incorporation of retrieval augmentation \citep{ren2023investigating}. More recently, \citet{xue2024ualign} utilize both semantic entropy and confidence signal to guide the behaviors of LLMs for known and unknown questions. Distinguished from previous works, our paper targets \textit{compositional questions}, considering various types of questions in practice.

\paragraph{Certainty and Uncertainty of LLMs.}
To calibrate the known and unknown of LLMs, there are lots of studies that have delved into methods for estimating and quantifying \textit{certainty and uncertainty} in LLMs predictions \citep{xiao-etal-2022-uncertainty, lin2022teaching, xiong2023llms, kuhn2023semantic}. There are two types of methods: 1) logit-based which utilize the model logits \citep{xiao-etal-2022-uncertainty, mielke-etal-2022-reducing}; and 2) non-logit-based methods, such as expressing uncertainty about its own answer in natural language \cite{lin2022teaching}, particularly with the rise of closed-source LLMs. More recently, \citet{xiong2023llms} benchmarks three categories of the first type: verbalize-based, consistency-based, and their hybrid methods. They find that LLMs exhibit a high degree of overconfidence when verbalizing their confidence, which can be alleviated by different prompting strategies (e.g., Chain-of-thoughts \citep{wei2023chainofthought}) or more complicated methods (e.g., Self-consistency \citep{wang2023selfconsistency}). Moreover, different languages also trigger different level of certainty and uncertainty of language models \cite{xue2024comprehensivestudymultilingualconfidence}.

\paragraph{Reasoning and Acting of LLMs.} On the one hand, lots of previous methods investigate various methods to elicit the internal reasoning capability of LLMs \citep{wei2023chainofthought, wang-etal-2023-cue, tpe}, such as program-guided reasoning \citep{pan-etal-2023-fact, khattab2023demonstratesearchpredict}, Self-Ask \citep{press2023measuring} and retrieval-augmented reasoning \citep{trivedi-etal-2023-interleaving, yu2023improving, shao-etal-2023-enhancing}, especially for multi-hop questions \citep{yang-etal-2018-hotpotqa} and in-depth dialogues \citep{wang-etal-2023-cue}. On the other hand, it is important to empower the stateless LLMs to interact with external world with the augmentation of different tools \citep{tool_tut}. Therefore, LLMs can perform tasks that go beyond their intrinsic knowledge such as retrieving up-to-date information \cite{wang-etal-2023-large, wang-etal-2024-uniretriever} and providing domain-specific services by calling different functions / APIs \cite{wang-etal-2024-appbench}. However, only a few of them consider the relationship between internal reasoning and external acting, especially for compositional problems when the necessary unknown knowledge is required. To address this dilemma, we explore the better trade-off between internal reasoning and external acting in terms of effectiveness and efficiency.

\section{Data Collection}
\label{sec:data_collection}
In this section, we thoroughly introduce how to collect the \textbf{C}ompositional \textbf{u}nknown \textbf{Q}uestion-\textbf{A}nswer dataset (CuQA) automatically, with the minimum human efforts to filter unqualified samples.

\subsection{Automatic Collection}
Algorithm~\ref{cuqa_algo} shows the pseudo-code details. Specifically, we assume there is a cutoff date for each LLM with the latest cutoff date for all LLMs, and all the pretraining corpus is collected before the cutoff date, for example, the cutoff date of \texttt{gpt4-turbo} is April 2023\footnote{\url{https://openai.com/blog/new-models-and-developer-products-announced-at-devday}}. In this way, we first collect all events that happened \textbf{after} the cutoff date from Wikipedia\footnote{\url{https://en.wikipedia.org/wiki/2023}}, named unknown events. Then we carefully implement different functions (i.e., UnknownQuestionGen) by prompting LLMs using different templates. We provide different information, e.g., internal known events and external unknown events, in the template to guide LLMs in generating the required output. For example, we use one entity in the unknown events as an answer and prompt the LLMs to generate corresponding questions according to the events (line 6). Appendix~\ref{cuqa_construction} shows the details of all functions' prompts. We finally store the questions, answers, and all intermediate results for further processing \footnote{It is worth noting that our data collection can be time-evolving given the cutoff date.}.

\begin{algorithm}[t]
\caption{CuQA Generation Algorithm}
    \label{cuqa_algo}
    \begin{algorithmic}[1]
    \REQUIRE Cutoff date $t$, Wikipedia $W$, LLM $\mathcal{M}$
    \ENSURE Generated Questions $\mathcal{Q}$
    \STATE $U_{e} = W (t)$  // get the unknown events according to the cutoff date of LLMs
    \FOR{$e_{j} \in U_{e}$}
    \STATE $ent_j$ = getEntities($e_{j}$) // get a list of entities
    \STATE $cur_{ent}$ = random.sample($ent_j$)
    \STATE $uk_{q}$ = UnknownQuestionGen($e_j, cur_{ent}$)
    \STATE $k_{e}$ = KnownEventsGen($cur_{ent}$)
    \IF{random.randint(1,9) < 5}
    \STATE $k_{q}$ = KnownQuestionGen($e_j, cur_{ent}$)
    \ELSE
    \STATE $k_{ent}$ = random.sample(getEntities($k_e$))
    \STATE $k_{q}$ = KnownQuestionGen2($k_e, k_{ent}, cur_{ent}$)
    \ENDIF
    \STATE $q$ = MergeQuestions($k_q, uk_q$)
    \STATE $\mathcal{Q}$.append(($q, cur_{ent}$ or $uk_q$))
    \ENDFOR
    \RETURN $\mathcal{Q}$
    \end{algorithmic}
\end{algorithm}

\subsection{Quality Control and Statistics}

To ensure the quality of the dataset, we additionally introduce some automatic quality control procedures and human evaluations. First of all, we write a Python script to validate whether or not the format of outputs meets the instructions in the functions. Moreover, we employ three well-educated annotators to: 1) filter unqualified samples ($\approx$10\%), such as answer is not correct or can not be inferred according to unknown events; and 2) rewrite the generated question to be more natural. Afterward, we successfully collect around 550 questions. It is worth noting 100 of them are hard questions which are further composed using multiple easy question-answering pairs (line 13). The examples in the data can be found in Appendix~\ref{data_cases}. 

\section{Method}
\begin{figure}
    \centering
    \includegraphics[trim={9cm, 5cm, 10cm, 3.5cm}, clip, width=0.48\textwidth]{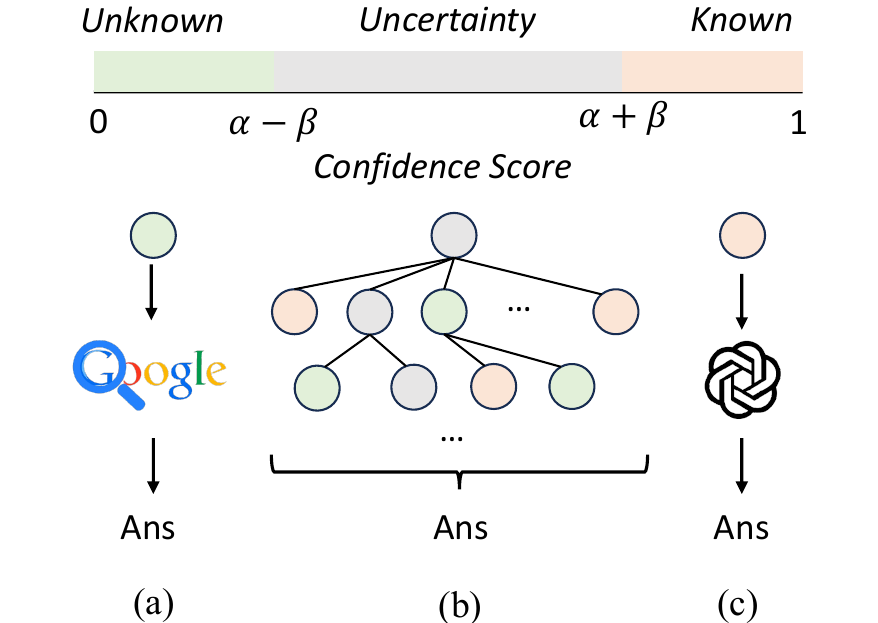}
    \caption{Overview of \textit{\texttt{Self-DC}}: a) retrieve-then-read for unknown questions, b) decompose-and-combination for uncertain questions; and c) generate-then-read for known questions.}
    \label{fig:algo}
    % \vspace{-4mm}
\end{figure}

To adaptively call different functions on-demand for \textit{compositional questions} understanding, it is essential to determine: \textbf{a)} whether the current question is known or unknown to the LLMs, and \textbf{b)} whether the current question can be further decomposed into different sub-questions. Therefore, given a question, we first get the confidence score of LLMs for a question and then (iteratively) call different functions, aiming to collect enough information to generate the final answer. Figure~\ref{fig:algo} shows an overview of the proposed self divide-and-conquer framework, \textit{\texttt{Self-DC}}.

% The confidence score and uncertainty zone ensure LLM can manage different types of questions by leveraging its strengths while mitigating its weaknesses.

% For the adaptive understanding of compositional questions, it is crucial to determine: a) whether the current question is known or unknown to the LLMs, and b) whether the current question can be further decomposed into different sub-questions. Consequently, we introduce a confidence score, denoted as $\alpha$, to represent an LLM's confidence in answering a question, along with a parameter, $\beta$, to indicate the uncertainty zone. The $\beta$ parameter assists the LLM in identifying when a question might be too complex or ambiguous for a direct answer, necessitating its decomposition into simpler parts or the amalgamation of multiple pieces of information. The confidence score and uncertainty zone ensure that the LLM can handle various types of questions by capitalizing on its strengths and mitigating its weaknesses.

\subsection{Framework: Self Divide-and-Conquer}

Since LLMs express certainty in different ways and are prone to hallucination issues, therefore, we define $\alpha$ as a mean of confidence score distribution for specific LLM, along with $\beta$ as the corresponding standard deviation. In this way, the LLMs can recognize when a question might be too complex or ambiguous for a straightforward answer, necessitating the decomposition into simpler parts or the combination of multiple pieces of information. Specifically, we divide the confidence score into three ranges $[0, \alpha-\beta], (\alpha-\beta, \alpha+\beta), [\alpha+\beta, 1]$. When the confidence score falls into extreme ranges, such as the left ($[0, \alpha-\beta]$) or right ($[\alpha+\beta, 1]$) side, we can directly apply retrieve-then-read or generate-then-read to answer the question respectively. However, when it encounters uncertain or confusing questions (i.e., fall into the middle part), we decompose the question into several sub-questions to decrease the uncertainty. We then iteratively solve these sub-questions in the same way and combine all sub-answers to answer the original \textit{compositional question} as shown in Figure~\ref{fig:algo_python}. To ensure efficiency and reduce unnecessary costs, we implement several pruning conditions to prevent iterations from overflowing: 1) the number of sub-questions is 1, which means it should be a \textit{Single Known} or \textit{Single Unknown} question; and 2) the number of iteration depth is less than a pre-defined $\tau$. Once these situations happen, we simply regard the current sub-question as the unknown question and then call retrieve-then-read. In this way, we can call compositional reasoning when necessary instead of treating all questions indiscriminately for different LLMs.

\subsection{Confidence Score Acquisition}
Inspired by lots of previous works \citep{lin2022teaching, xiong2023llms}, we use two types of method to prompt the LLM itself to get the confidence score to answer the question.

\begin{itemize}[leftmargin=*]
    \item \textit{\textbf{verbalize-based} (verb).} We instruct the LLMs to output the confidence level from 0 to 100 following the answer to the question \citep{xiong2023llms}. We clearly note that the confidence level indicates the degree of certainty. Then we re-map the confidence score to the range $[0,1]$. The details of the prompt can be found in Appendix.
    
    \item \textit{\textbf{probability-based} (prob).} We additionally utilize the probability information to calculate the confidence score. Specifically, we firstly prompt the LLMs to generate the answer using a few words, 
    and then we get the probability $\hat p_i$ of $i$-th token in the generated content. 
    % and then we get the log-probabilities of each token in the generated content. 
    We take the average of probabilities in the sequence as the confidence score \cite{varshney2023stitch} following Eq.~\ref{eq_avg}:
    
    \begin{equation}
    \label{eq_avg}
        conf = \frac{1}{N} \sum_{i=1}^{N} \hat{p}_i
    \end{equation}

\end{itemize}

\noindent Considering the poor performance of LLMs to express uncertainty as reported by lots of existing works \citep{lin2022teaching, xiong2023llms} and complex situations in practice, we additionally introduce $\alpha$ and $\beta$ to control the range of uncertainty, enhancing the flexibility and robustness of \textit{\texttt{Self-DC}}.

\subsection{Other Sub-Functions}
According to different levels of confidence scores, we carefully design several functions to complete the compositional reasoning task, aiming to provide a more accurate answer. We present the details of other sub-functions one by one as follows:

\begin{itemize}[leftmargin=*]
    \item \textbf{Generate-then-read:} Following \citet{genread}, we firstly prompt the LLM to generate a background document from Wikipedia to answer the given question, and then ask the LLM to answer the question by referring to the generated passage. The prompt details can be found in the original paper.

    \item \textbf{Retrieve-then-read:} We utilize the retriever to retrieve external knowledge at the first step and then ask the LLM to answer the question by referring to the retrieved passage.

    \item \textbf{Decompose:} We prompt the LLMs to systematically break down the overarching question into several smaller sub-questions. The answers to these sub-questions collectively contribute to deriving the answer to the original overarching question, similar to \citet{press2023measuring} and \citet{xu2023searchinthechain}.

    \item \textbf{Combine answers:} After the decomposition, we call the main function to enter the next iteration as shown in Figure~\ref{fig:algo_python}, aiming to get the answer to each sub-question. Subsequently, we combine the answers to all sub-questions to get the answer to the original question.
\end{itemize}

\lstset{
  backgroundcolor=\color{white},
  basicstyle=\fontsize{7.5pt}{8.5pt}\fontfamily{lmtt}\selectfont,
  columns=fullflexible,
  breaklines=true,
  captionpos=b,
  commentstyle=\fontsize{8pt}{9pt}\color{codegray},
  keywordstyle=\fontsize{8pt}{9pt}\color{codegreen},
  stringstyle=\fontsize{8pt}{9pt}\color{codeblue},
  frame=tb,
  emph={decompose, combine_sub_qas, generate_then_read, retrieve_then_read, SelfDC, get_confidence_score},
  emphstyle={\bfseries\color{orange}}
}
\begin{figure}[t]
\tiny
\begin{lstlisting}[language=python]
def SelfDC(m, r, q, alpha, beta):
    # m: large language model
    # r: retriever for searching documents
    # q: question to be answered
    # alpha, beta: hyperparameters for defining ranges

    c = get_confidence_score(m, q)

    if c < alpha + beta and c > alpha - beta:
        sub_qs = decompose(m, q)
        sub_as = [SelfDC(m, r, sub_q, alpha, beta) for sub_q in sub_qs]
        answer = combine_sub_qas(m, q, sub_qs, sub_as)
    elif c >= alpha + beta:
        answer = generate_then_read(m, q)
    else:
        answer = retrieve_then_read(m, r, q)

    return answer
\end{lstlisting}
\caption{The simplified python implementation details of \texttt{Self-DC}, consisting of several functions: 1) \textit{decompose}; 2) \textit{combine-sub-qas}; 3) \textit{generate-then-read}; and 4) \textit{retrieve-then-read}.}
\label{fig:algo_python}
\vspace{-4mm}
\end{figure}

\section{Experiment}

\subsection{Baselines and Evaluation Metrics}
\paragraph{Baselines.} To provide a comprehensive evaluation, we compare our method with different prompting methods with or without the involvement of retrieval augmentation: 1) \textbf{Direct Prompting} \citep{brown2020language}; 2) \textbf{Chain-of-thought (CoT) prompting} \citep{wei2023chainofthought}, including zero-shot and few-shot setting; 3) \textbf{GenRead} \citep{genread} which firstly prompts the LLMs to generate known knowledge and then answer the question; 4) \textbf{Retrieve-then-read (RR)} which retrieves the related passages first and then answers the questions, following \citet{yu2023improving}; 5) \textbf{Self-Ask} \citep{press2023measuring} involves generating follow-up questions, retrieving information based on those questions, and providing answers, until no more follow-up questions are generated and the LLMs answer the original question at the last; 6) \textbf{IRCoT} \citep{trivedi-etal-2023-interleaving} interleaves retrieval with steps (sentences) in a CoT, guiding the retrieval with CoT and in turn using retrieved results to improve CoT; 7) \textbf{REFEED} \citep{yu2023improving} and 8) \textbf{ITER-RETGEN} \citep{shao-etal-2023-enhancing} utilize the generated answer or intermediate reasoning results to enrich the query, leading to better retrieval and final answer to original question, respectively.

\paragraph{Datasets and Evaluation Metrics.} We conduct our experiments mainly on two datasets: 1) the newly proposed CuQA dataset; and 2) FreshQA \citep{vu2023freshllms}, which contains 600 question-answer pairs that require fast-changing world knowledge, including the latest ones \footnote{We use the version on 30th Sep, 2024.}. We note here that FreshQA is not a typical compositional QA dataset despite it containing few \textit{compositional questions}. To select suitable values for $\alpha$ and $\beta$, we randomly sample 50 instances as a development set for CuQA, leaving 500 instances for testing. For FreshQA, we use the original split: 500 test instances and 100 development instances. Following previous works \citep{genread, yu2023improving, trivedi-etal-2023-interleaving}, we select Exact Match (EM)\footnote{We consider it is matched when the predicted answer in the ground truth answer due to various outputs by LLMs.}, F1 to evaluate the performance of different methods. Furthermore, to enhance the robustness of the evaluation, we use Acc$^\dagger$ as an additional metric and prompt LLMs to assess the predictions related to the actual ground-truth answers following \citet{shao-etal-2023-enhancing}.

\subsection{Implementation Details}
We mainly conduct our experiments on two different backbone models: \texttt{gpt-3.5-turbo-1106} and \texttt{gpt-4o-mini}, hereinafter referred to as \texttt{1106} and \texttt{4o-mini} respectively, following lots of previous works \citep{genread,yu2023improving,shao-etal-2023-enhancing}. For the Acc$^\dagger$ evaluation, we always use \texttt{4o-mini} as evaluation backbone model. We set both the temperature and top p as 0.1 to reduce the randomness of LLMs for all methods, rendering a more fair comparison. We implement the Google search engine following \texttt{LangChain} \footnote{\url{https://python.langchain.com/docs/integrations/tools/google_search}} as an external retriever, and we set the number of retrieved results as 3 and the max iteration depth $\tau$ as 3. According to the preliminary results on the validation set, we fix $\beta$ as 0.1 and $\alpha$ as 0.9 for verb (0.8 for prob) on \texttt{1106} for both datasets, and $\alpha$ as 0.6 for verb (0.6 for prob on CuQA; 0.8 for prob on FreshQA) on \texttt{4o-mini}. The significant test (t-test) is conducted with $p$ < 0.05 to ensure statistical improvement.

\begin{table}[!t]
  \centering
  \begin{adjustbox}{max width=0.48\textwidth}
  \begin{tabular}{l|c|ccc|ccc}
    \toprule
    \multirow{2}{*}{\textbf{Methods}} & \multirow{2}{*}{\#\textbf{R}} & \multicolumn{3}{c|}{\textbf{CuQA}} & \multicolumn{3}{c}{\textbf{FreshQA}} \\
    
    \cline{3-8} & & EM & F1 & Acc$^\dagger$ & EM & F1 & Acc$^\dagger$  \\
    \hline
    \hline
    \multicolumn{7}{c}{\textbf{\textit{w/o retrieval}}} \\
    \hline
    \hline
    Direct & 0 & 21.0 & 19.3 & 34.2 & 20.6 & 21.6 & 37.6 \\
    CoT & 0 & 21.8 & 20.5 & 36.6 & 21.2 & 22.9 & 38.8  \\
    Few-shot-CoT$^*$ & 0 & 7.2 & 1.7 & 9.6 & 18.0 & 11.1 & 26.8 \\
    GenRead & 0 & 12.2 & 12.6 & 23.2 & 18.8 & 19.3 & 36.0 \\
    \hline
    \hline
    \multicolumn{7}{c}{\textbf{\textit{w/ retrieval}}} \\
    \hline
    \hline
    RR & $n$ & 30.4 & \textbf{24.7} & 48.2 & 34.2 & \textbf{28.9} & \underline{61.6} \\
    REFEED & $2n$ & \underline{35.2} & 8.2 & \textbf{53.2} & 29.6 & 16.1 & 49.2 \\
    IRCoT & $3n$ & \textbf{39.0} & 8.1 & 50.4 & 32.0 & 15.5 & 61.2  \\
    Self-Ask$^*$ & $0$-$n$ & 8.6 & 4.3 & 11.2 & 16.8 & 13.4 & 27.4 \\
    ITER-RETGEN$^*$ & $2n$ & 19.2 & 5.8 & 25.4 & 32.4 & 15.7 & 46.6 \\
    \hline
    \textit{\textbf{Self-DC}} (\textit{verb}) & $0$-$2n$ & 31.8 & 20.4 & 49.4 & \underline{34.3} & 25.2 & 58.1 \\
    \textit{\textbf{Self-DC}} (\textit{prob}) & $0$-$n$ & 32.6 & \underline{21.7} & \underline{50.6} & \textbf{36.2} & \underline{28.4} & \textbf{62.2} \\
    \bottomrule
\end{tabular}
\end{adjustbox}
\caption{The performance of baselines and \texttt{Self-DC} with the \texttt{1106}. The baseline$^*$ means it uses demonstrations and The column \#\textbf{R} denotes the number of retrieval calls in terms of number of test cases $n$. We \textbf{bold} the best performance and \underline{underline} the second-best performance.}
% \vspace{-4mm}
\label{tab:main_exp_1106}
\end{table}

\subsection{Main Results}

Table~\ref{tab:main_exp_1106} and Table~\ref{tab:main_exp_4o_mini} show the performances of all baselines and our proposed \textit{\texttt{Self-DC}} on the \texttt{1106} and \texttt{4o-mini} respectively. Therefore, several conclusions can be drawn from the results:

\paragraph{\textit{CoT (or Few-shot-CoT) does not bring consistent improvements over direct prompting (Direct).}} We surprisingly found that the performance of CoT at both Table~\ref{tab:main_exp_1106} and Table~\ref{tab:main_exp_4o_mini} is usually worse than Direct, and Few-shot-CoT can not further boost the performance particularly with \texttt{1106}, revealing the complexity of compositional reasoning.

\paragraph{\textit{Retrieval-based method generally achieves better performance than non-retrieval methods but the gap is smaller with \textit{compositional questions}}.} It is observed that RR and IRCoT are capable of achieving better performance than non-retrieval baselines, and IRCoT sometimes achieves the highest performance due to a more complex retrieval design, accompanied by more cost. Secondly, the gap between retrieval-based and non-retrieval-based methods on FreshQA is relatively larger than on CuQA. This discrepancy is likely because CuQA contains more \textit{compositional questions}, which, when used directly as queries, result in noisier documents. Furthermore, we surprisingly observe that Self-Ask and ITER-RETGEN achieve the lowest performance, especially on CuQA. To understand the reason, we examined the intermediate reasoning results and found that Self-Ask tends not to generate follow-up questions and directly answer the question, rarely calling for retrieval given the compositional unknown question. On the other hand, ITER-RETGEN retrieves external documents step-by-step but introduces a lot of noise since the queries are mostly related to the original \textit{compositional question}. These observations reveal the significance and valuable insights provided by the CuQA dataset, highlighting its importance for understanding the challenges associated with \textit{compositional questions}.

\begin{table}[!t]
  \centering
  \begin{adjustbox}{max width=0.48\textwidth}
  \begin{tabular}{l|c|ccc|ccc}
    \toprule
    \multirow{2}{*}{\textbf{Methods}} & \multirow{2}{*}{\#\textbf{R}} & \multicolumn{3}{c|}{\textbf{CuQA}} & \multicolumn{3}{c}{\textbf{FreshQA}} \\
    
    \cline{3-8} & & EM & F1 & Acc$^\dagger$ & EM & F1 & Acc$^\dagger$  \\
    \hline
    \hline
    \multicolumn{7}{c}{\textbf{\textit{w/o retrieval}}} \\
    \hline
    \hline
    Direct & 0 & 29.0 & 19.4 & 46.4 & 27.2 & 17.3 & 53.0 \\
    CoT & 0 & 28.8 & 18.2 & 46.0 & 29.2 & 18.1 & 53.8 \\
    Few-shot-CoT$^*$ & 0 & 43.0 & 3.2 & 50.8 & 35.0 & 9.1 & 55.4 \\
    GenRead & 0 & 29.6 & 29.2 & 47.4 & 26.8 & 27.7 & 52.0 \\
    \hline
    \hline
    \multicolumn{7}{c}{\textbf{\textit{w/ retrieval}}} \\
    \hline
    \hline
    RR & $n$ & 32.0 & 31.6 & 55.4 & \underline{35.2} & 32.6 & \underline{63.4} \\
    REFEED & $2n$ & 26.2 & \underline{33.5} & 51.8 & 28.8 & \underline{34.5} & 57.4 \\
    IRCoT & $3n$ & \textbf{47.8} & 13.5 & \textbf{64.6} & 34.2 & 17.8 & 61.4 \\
    Self-Ask$^*$ & $0$-$n$ & 19.8 & 3.8 & 48.4 & 5.6 & 9.8 & 59.0 \\
    ITER-RETGEN$^*$ & $2n$ & 23.4 & 12.6 & 50.9 & 31.2 & 21.1 & 55.8  \\
    \hline
    \textit{\textbf{Self-DC}} (\textit{verb}) & $0$-$n$ & 34.0 & 32.2 & 53.8 & 30.2 & 30.2 & 59.8 \\
    \textit{\textbf{Self-DC}} (\textit{prob}) & $0$-$n$ & \underline{36.4} & \textbf{36.5} & \underline{56.4} & \textbf{37.4} & \textbf{36.6} & \textbf{66.4} \\
    \bottomrule
\end{tabular}
\end{adjustbox}
\caption{The performance of baselines and \texttt{Self-DC} with the \texttt{4o-mini}. }
\vspace{-4mm}
\label{tab:main_exp_4o_mini}
\end{table}

\paragraph{\textit{\texttt{Self-DC}} achieves better trade-off between efficiency and effectiveness than retrieval-based methods.} When comparing \textit{\texttt{Self-DC}} to other baselines considering the consumption of retrieval calls (\#R), it is evident that \textit{\texttt{Self-DC}} achieves better performance compares with the method utilizing same or more calls, for example, \textit{\texttt{Self-DC}} (prob) v.s. RR. Even compared with some methods that require 2 to 3 times more retrieval, \textit{\texttt{Self-DC}} still achieves comparable results and even outperforms them in specific dataset. This is important to highlight, as it not only establishes an effective and efficient framework to call external retrieval, but also demonstrates a promising path for controlling the behavior of LLMs by leveraging the internal signals they generate (i.e., the internal confidence scores).

\begin{figure}[t]
    \centering
    \includegraphics[width=0.48\textwidth]{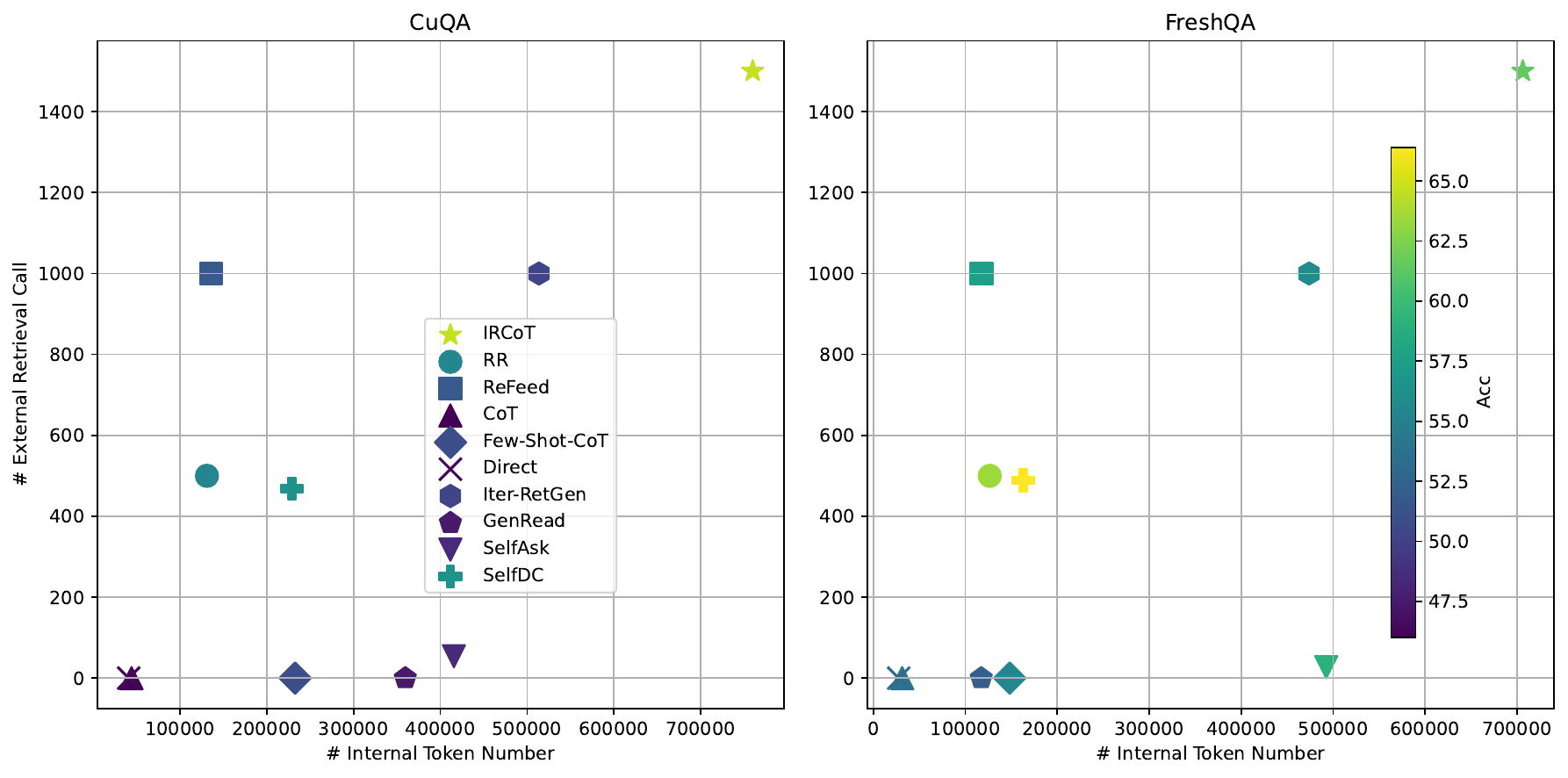}
    \caption{The efficiency analysis of different methods using \texttt{4o-mini}.}
    \label{fig:efficiency_ana_4o}
    % \vspace{-6mm}
\end{figure}

\section{Analysis}
In this section, we present a comprehensive analysis of \textit{\texttt{Self-DC}} mainly using the CuQA dataset, covering three key aspects: efficiency analysis, the choices of $\alpha$ and $\beta$ and different iteration depth on latest model \texttt{gpt-4o-mini}.

\subsection{Efficiency Analysis}
\label{sec:efficient_ana}

To directly validate the efficiency of \textit{\texttt{Self-DC}}, we consider three dimensions: \# internal token consumption, \# external retrieval calls and the final performance. Table~\ref{fig:efficiency_ana_4o} illustrate the report. Ideally, we aim for a method which achieves the best performance appears at the left bottom of figure. Only in such a case, the method would demonstrate its superiority by not only delivering better performance but, more importantly, by eliciting the great potential of the internal capabilities of LLMs and minimizing reliance on external resources or tools. According to the figure, it is obvious that \textit{\texttt{Self-DC}} achieves great balance between these three factors. It is worthy noting we observe similar trends on \texttt{1106} for both datasets.

\subsection{The Impacts of Different $\alpha$ and $\beta$}
\begin{figure}[t]
    \centering
    \includegraphics[width=0.48\textwidth]{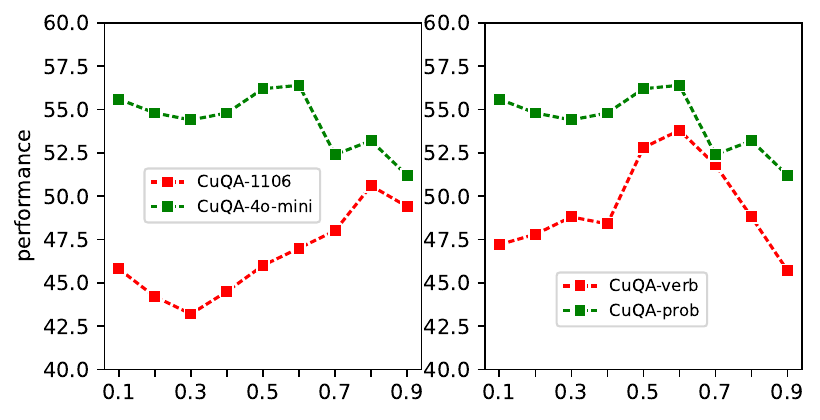}
    \caption{The performance of different choices of $\alpha$ with $\beta$ = 0.1. \textbf{Left:} The performance of different models with confidence type is \textit{prob}; and \textbf{Right:} The performance of different confidence types (\textit{verb} or \textit{prob}) with the same model \texttt{4o-mini}.}
    \label{fig:alpha_impacts_4o}
\end{figure}

It is vital to balance alpha and beta for optimizing the performance of LLMs to different tasks. In this section, we provide detailed analysis of different choices of $\alpha$ and $\beta$. Firstly, we fix $\beta=0.1$ and set $\alpha$ to $[0.1, 0.2, 0.3, ..., 0.9]$. The results can be found in Figure~\ref{fig:alpha_impacts_4o}. The entire processing can be seen as a 0.2-length uncertainty block starts from 0 to 1 with stride = 0.1. First of all, We found that none of the lines shows monotonically increasing or decreasing, and most of the best performances are achieved in the middle choice of $\alpha$, revealing the complexity of the target problem. In detail, there is an upward and then downward trend globally (e.g., in the right figure). It is reasonable since LLMs utilize more generate-then-read functions at the beginning (e.g., $\alpha$=0.1, $\beta$=0.1), resulting in poor performance. With the uncertainty, blocks move to the right side (a.k.a, 1), LLMs will utilize retrieve-then-read more frequently. Once exceeds a specific threshold, the performance will drop since the decomposition will introduce more noise compared with gains. 

\begin{figure}[t]
    \centering
    \includegraphics[width=0.48\textwidth]{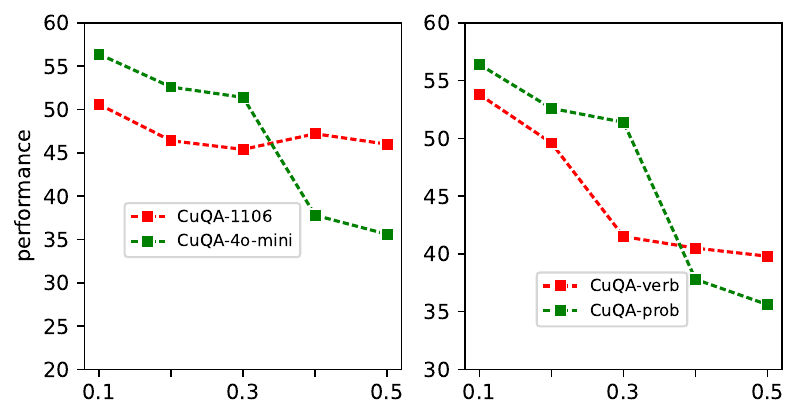}
    \caption{The performance of different choices of $\beta$ with a fixed $\alpha$ as 0.8 for \texttt{1106} and 0.6 for \texttt{4o-mini}. \textbf{Left:} The performance of different models with confidence type is \textit{prob}; and \textbf{Right:} The performance of different confidence types (\textit{verb} or \textit{prob}) with the same model \texttt{4o-mini}.}
    % \vspace{-6mm}
    \label{fig:beta_impacts_4o}
\end{figure}

Secondly, we fix $\alpha$ with different values according to the best performance above and set $\beta$ to $[0.1, 0.2, ..., 0.5]$ to investigate the impacts of different $\beta$. Figure~\ref{fig:beta_impacts_4o} shows the final results. It is obvious that there is a monotonically decreasing trend. After carefully checking the specific confidence scores distributions, we attribute this to be smaller range changes in the score. In general, despite that the choices of $\alpha$ and $\beta$ are extremely tricky with lots of factors in practice, we humbly point out that most of simply combination (e.g., $\alpha$=0.5, $\beta$=0.1) achieves comparable performance with baselines require more retrieval or token consumption even it may not be optimal combination.

\subsection{The Impacts of Different Iteration Depth}

Table~\ref{tab:iteration_times} shows the results. First of all, we can find that different choices of $\tau$ have a slight effects on the final performance. As the iteration depth increases, the number of retrieval calls rises correspondingly, as noted in prob, while verb remains largely unchanged. We suspect this is due to \textit{verb} is not as accurate as \textit{prob}. In this way, it calls almost all external retrieval for unknown questions only within the shallow iteration. Most importantly, we want to emphasise here that the number of retrieval calls usually will not exceed the number of original test set $n$, and sometimes it only need to call less than $0.5n$ calls, revealing the great advantages of \textit{\texttt{Self-DC}} over other iterative retrieval-augmented baselines. 

\begin{table}[!t]
  \centering
  \begin{adjustbox}{max width=0.48\textwidth}
  \begin{tabular}{l|c|c|c}
    \toprule
    \textbf{Model} & \textbf{2} & \textbf{3} & \textbf{4} \\
    \hline
    \texttt{4o-mini} (\textit{verb}) & 50.2 (76) & 53.8 (78) & 53.4 (78) \\
    \texttt{4o-mini} (\textit{prob}) & 52.4 (455) & 56.4 (468) & 55.3 (470) \\
    \bottomrule
\end{tabular}
\end{adjustbox}
\caption{The performance of \texttt{Self-DC} with different max iteration times. We also report the number of retrieval times in the (bracket).}
\label{tab:iteration_times}
% \vspace{-6mm}
\end{table}

\subsection{Error Analysis}
\label{sec:error-analysis}

\paragraph{Performance of different types of questions.} Table~\ref{tab:types_of_questions} shows the results of different types of questions in CuQA. There exists a significant disparity in performance between easy and difficult questions, indicating a substantial challenge for models when addressing complex compositional unknown questions. Upon analyzing the error cases, we identified several prevalent issues: 40\% of errors arise from repetitive sub-questions, 13\% are due to irrelevant or incorrect sub-questions, such as "\textit{What month is it now?}", another 13\% involve correct decomposition but incorrect answers.

\begin{table}[!t]
  \centering
  \begin{adjustbox}{max width=0.35\textwidth}
  \begin{tabular}{l|c|c|c}
    \toprule
    \textbf{Types} & \hspace{2mm} \textbf{EM} \hspace{2mm} & \hspace{2mm} \textbf{F1} \hspace{2mm} & \hspace{2mm} \textbf{Acc}$^\dagger$ \hspace{2mm} \\
    \hline
    Easy & 38.1 & 37.6 & 58.8 \\
    Hard & 26.7 & 22.7 & 33.3 \\
    \bottomrule
\end{tabular}
\end{adjustbox}
\caption{The performance of \texttt{Self-DC} on two types of question: \textit{easy} and \textit{hard} in CuQA using \texttt{4o-mini}.}
\label{tab:types_of_questions}
% \vspace{-5mm}
\end{table}

\paragraph{Accuracy of confidence scores.} First of all, when using \textit{verb} method, we find that the confidence scores are 0 for more than 65\% cases, and over 0.9 for around 20\% cases with \texttt{1106}.
However, the trend is slightly different when it comes to \texttt{4o-mini} which gives 0.9 more frequently ($\approx$ 35\%). These two scores represent the top two most frequently occurring scores in both models. It seems LLMs either overestimate the correctness, or directly acknowledge the uncertainty and refuse to answer. Moreover, there is pretty rare of fine-grained confidence score (i.e, 0.82, 0.61), making the fine-grained choices of $\beta$ meaningless in \textit{verb}. On the other hand, when using \textit{prob} method, there are much more fine-grained confidence signals, and most of them falls in the $< 0.5$ part ($\approx$ 90\%). It is clear that \textit{prob} leads to better performance compared with \textit{verb} and generally \texttt{4o-mini} outperforms \texttt{1106}.

\begin{figure*}[h]
    \includegraphics[trim={3cm, 6cm, 4cm, 3cm}, clip, width=0.9\textwidth]{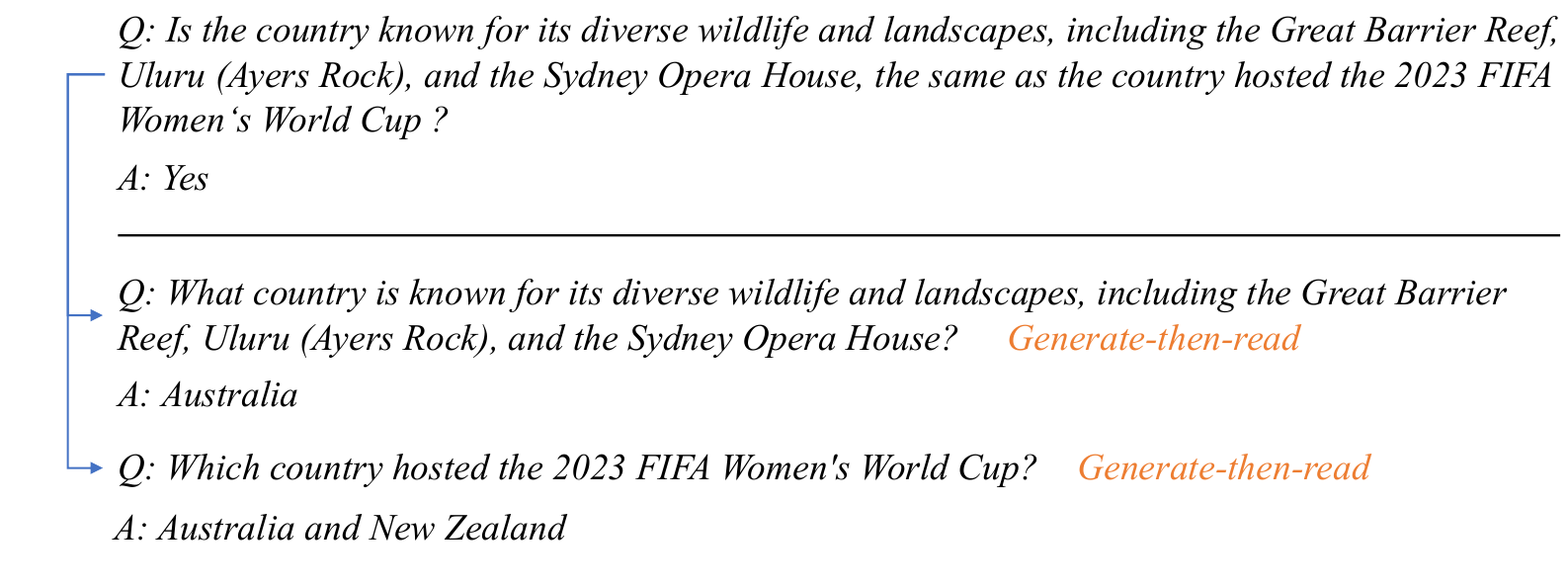}
    \caption{An example from CuQA dataset where one \textit{compositional  question} can be further divided into two known sub-questions.}
    \label{cuqa_case_study}
\end{figure*}

\begin{figure*}[h]
    \includegraphics[trim={3cm, 5cm, 4cm, 3cm}, clip, width=0.9\textwidth]{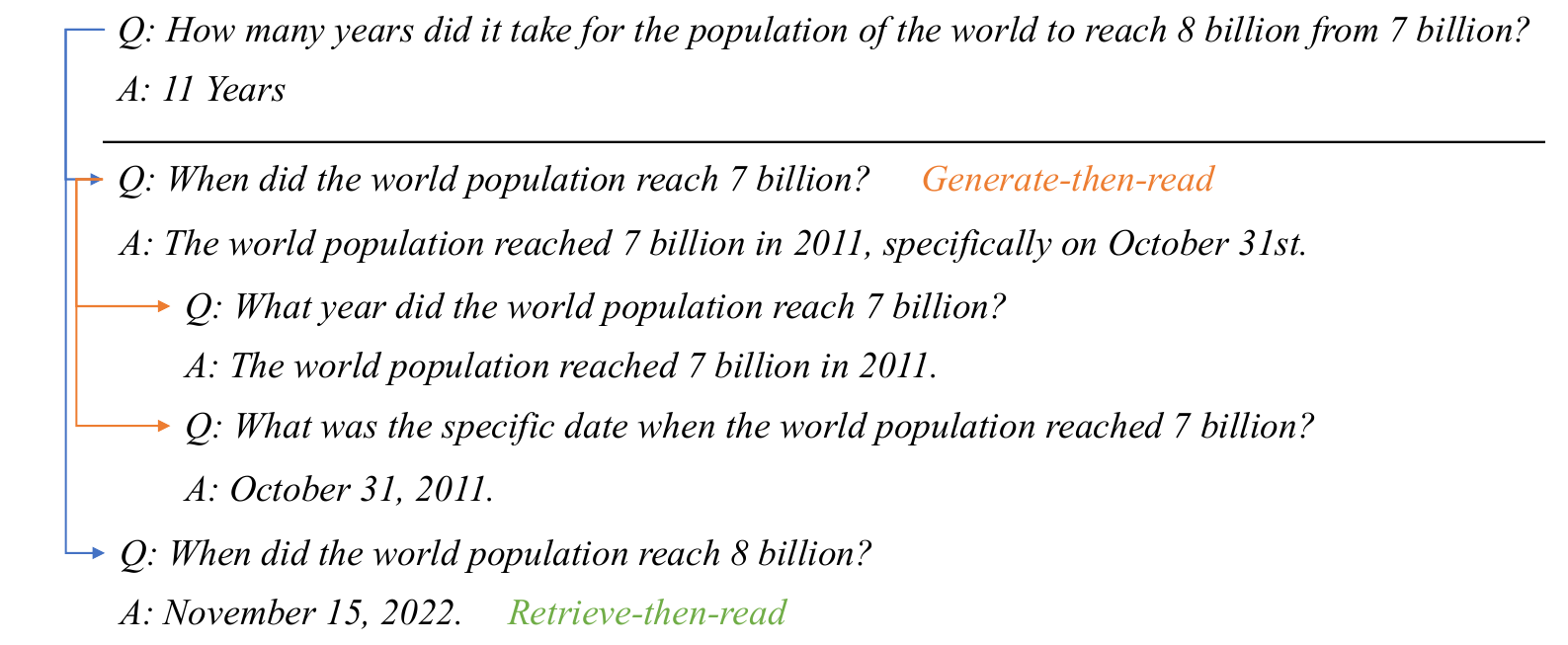}
    \caption{An example from FreshQA dataset where one \textit{compositional question} can be further divided into known and unknown questions.}
    \label{freshqa_case_study}
\end{figure*}

\paragraph{Analysis of decomposition} 
The times of decomposition are highly affected by the confidence scores. Sometimes, the best performance can be achieved without any decomposition with well-selected $\alpha$ and $\beta$. We study the percentage of decomposition and how many original questions are answered correctly after decomposition. We found that 0\% (i.e., $\alpha$=0.9) to 80\% (i.e., $\alpha$=0.1) questions will be decomposed and it is about 40\% to 50\% questions are correctly answered after decomposition \footnote{The case study and more analysis can be found in Appendix.}.

\section{Case Study}
Figure~\ref{cuqa_case_study} and Figure~\ref{freshqa_case_study} show an example from CuQA and FreshQA dataset respectively. We can found that \textit{\texttt{Self-DC}} is capable to call different functions to address various \textit{compositional questions} including known and unknown sub-questions.

\section{Conclusion}
In this paper, we firstly introduce compositional unknown questions, which contain several known and unknown sub-questions. We build a benchmark, named CuQA, to evaluate the performance and efficiency of existing compositional reasoning methods. Furthermore, we present a Self Divide-and-Conquer (\textit{\texttt{Self-DC}}) method to adaptively call external or internal knowledge, which not only demonstrates comparable or even better performance compared with existing complex iterative retrieval methods with fewer retrieval calls but also shows a promising potential to elicit internal capabilities of LLMs while minimizing external reliance.

\section*{Limitations}
\label{sec:limitations}

We discuss two major limitations in this paper regarding the dataset and method issues.

\paragraph{Dataset and Model.} Due to space limitations and cost, we choose to conduct our experiments on two datasets and two models. We would like to evaluate the performance of more models, i.e., several open-source models, on the proposed datasets and more \textit{compositional questions}.

\paragraph{Method.} We mainly implement our method in zero-shot setting, and do not consider more complex implementation for each function within the framework, in order to demonstrate the great potential and effectiveness of our proposed method more clearly and straightforwardly. We left more complex implementations in our future works. 

Furthermore, we would like to discuss the hallucination issues or other issues from the model side. Since different LLMs express certainty in various levels and may hallucinate the confidence score, we have meticulously designed the parameters $\alpha$ and $\beta$ to ensure that our framework remains flexible and easily adaptable to a broader range of LLMs. While we acknowledge it may be relatively difficult to choose them, we are encouraged to see more and more recent studies align certainty expression across LLMs \cite{tao2024trust, xu2024sayself, lee2024improving} and our method still outperforms other baselines even with the existing of these issues. From a dynamic and development standpoint, we believe our method and dataset could play a key role in the field of \textit{compositional question} answering. 

\section*{Ethical Statements}
In this paper, there are only one issue about dataset collection.

\paragraph{Human Annotation} The human inspection and annotation process are conducted by a respected data annotation company. All annotators receive fair compensation based on market rates and their personal information is not disclosed.

\section*{Acknowledgement}

This work was partially supported by Hong Kong RGC GRF No. 14206324, CUHK direct grant No. 4055209, and CUHK Knowledge Transfer Project Fund No. KPF23GWP20.

\bibliography{custom}
\bibliographystyle{acl_natbib}
\clearpage

% \begin{figure*}[t]
%     \centering
%     \includegraphics[width=0.99\textwidth]{figs/data_build.png}
%     \caption{Data Construction.}
%     \label{fig:incon}
% \end{figure*}

\appendix
\section{Data Collection}
\label{data_collection}

% \subsection{Prompt Templates}   
% \label{prompt_templates}

% \input{templates/confidence_scores}

\subsection{CuQA Construction}
\label{cuqa_construction}
We detail the prompts used for dataset construction in Tables \ref{tab:data-collection-unknow-quesgen}-\ref{tab:data-collection-mergequesgen}.
\begin{table}[H]
\small
    \centering
    \colorbox{blue!8}{
    \begin{tabular}{@{}p{7.2cm}}
    Unknown: \{\texttt{unknown\_event}\}
    
    According to the unknown event, please generate a question to which the answer is the entity \{\texttt{unknown\_entity}\}.
    \end{tabular}
    }
    \caption{\texttt{UnknownQuestionGen()} function in Algorithm \ref{cuqa_algo}: generate an \texttt{unknown\_question} about the \texttt{unknown\_event} with the \texttt{unknown\_entity} serving as the answer.
    % Unknown question generation of the selected entity from unknown events.
    }
    \label{tab:data-collection-unknow-quesgen}
\end{table}
% \begin{table}[ht]
% \small
%     \centering
%     \colorbox{blue!8}{
%     \begin{tabular}{@{}p{7.2cm}}
%     Unseen: \{event\}
    
%     Question: \{question\} 
    
%     Referring to the information above, the correct answer in just one entity to the given question is
%     \end{tabular}
%     }
%     \caption{Answer generation of the generated question with entity from unknown events for quality control. \{question\} is the output of Table \ref{tab:data-collection-unknow-quesgen}.}
%     \label{tab:data-collection-unknow-answergen}
% \end{table}
\begin{table}[H]
\small
    \centering
    \colorbox{blue!8}{
    \begin{tabular}{@{}p{7.2cm}}
    Generate a detailed passage about \{\texttt{entity}\}
    \end{tabular}
    }
    \caption{\texttt{KnownEventsGen()}
    function in Algorithm \ref{cuqa_algo}: generate a supporting background information \texttt{known\_events} about the \texttt{unknown\_entity} based on the internal known knowledge of LLMs.
    % supporting background information generation of the selected entity from unknown events based on the parametric knowledge of LLMs.
    }
    \label{tab:data-collection-known-docgen}
\end{table}

\begin{table}[H]
\small
    \centering
    \colorbox{blue!8}{
    \begin{tabular}{@{}p{7.2cm}}
    Known: \{\texttt{known\_event}\}
    
    According to known events, please generate a question to which the answer is be the entity \{\texttt{entity}\}.
    \end{tabular}
    }
    \caption{\texttt{KnownQuestionGen()} function in Algorithm \ref{cuqa_algo}: generate a \texttt{known\_question} based on the \texttt{known\_event} with the \texttt{entity} serving as the answer.
    % Question generation of the selected entity from unknown events based on the generated passage from Table \ref{tab:data-collection-known-docgen}.
    }
    \label{tab:data-collection-known-quesgen-AAB}
\end{table}

\begin{table}[H]
\small
    \centering
    \colorbox{blue!8}{
    \begin{tabular}{@{}p{7.2cm}}
    Seen: \{\texttt{known\_passage}\}
    
    Generate a question that meets the following conditions: 1. contains the terms \{\texttt{unknown\_entity}\} in question, 2. the answer is \{\texttt{known\_entity}\}.
    \end{tabular}
    }
    \caption{\texttt{KnownQuestionGen2()} function in Algorithm \ref{cuqa_algo}: given the  \texttt{known\_event}, generate a \texttt{known\_question} which contains the \texttt{unknown\_entity} in the question and can be answered with \texttt{known\_entity} sampled from the \texttt{known\_event}.
    }
    \label{tab:data-collection-known-quesgen-ABC}
\end{table}

\begin{table}[H]
\small
    \centering
    \colorbox{blue!8}{
    \begin{tabular}{@{}p{7.2cm}}
    Question One: \{\texttt{unknown\_question}\} 
    
    Question Two: \{\texttt{known\_question}\} 
    
    Generate a more natural combined question of question one and question two.
    \end{tabular}
    }
    \caption{\texttt{MergeQuestions()} function in Algorithm \ref{cuqa_algo}: merge the generated \texttt{unknown\_question} and \texttt{known\_question} into a single multi-hop question.
    % merge the unknown question from Table \ref{tab:data-collection-unknow-quesgen} and known question from \ref{tab:data-collection-known-quesgen} into a single one.
    }
    \label{tab:data-collection-mergequesgen}
\end{table}

\subsection{Data Examples}
We list two easy examples from CuQA dataset in Table \ref{tab:data_examples}. There are two reasoning types in CuQA: (1) \texttt{AAB} represents the two questions \texttt{Q1} and \texttt{Q2} are independently created before being merged; (2) \texttt{ABC} means the generation of \texttt{Q2} depends on \texttt{Q1}, where in the listed example, \texttt{A1} is embedded within \texttt{Q2}. It means the three QA pairs are synthesized in a concatenated form. We also regard two \texttt{merged} QA pairs as the sub-problems, combining them to form a more complex question that demands enhanced reasoning and more decomposition.
\label{data_cases}
\begin{table*}[t!]
\centering
\begin{adjustbox}{max width=0.95\textwidth}
\begin{tabular}{c|l}
\toprule
\multicolumn{1}{c|}{\textbf{Reasoning Type}} & \multicolumn{1}{c}{\textbf{Examples}} \\ 
\midrule
AAB                       & \begin{tabular}[c]{@{}l@{}}
\textbf{Q1:} Which countries signed a trilateral pact on 18 August, 2023?\\ 
\textbf{A1:} The United States, Japan, and South Korea\\ \\ 
\textbf{Q2:} What's the G7 member countries?\\ 
\textbf{A2:} Canada, France, Germany, Italy, Japan, the United Kingdom and the United States.\\ \\ 
\textbf{Merged-Q:} Which two G7 member countries signed a trilateral pact on 18 August, 2023?\\ 
\textbf{Merged-A:} The United States, Japan.\end{tabular} \\ 
\midrule

ABC                       & \begin{tabular}[c]{@{}l@{}}
\textbf{Q1:} Where did the first AI Safety Summit take place?\\ 
\textbf{A1:} United Kingdom\\ \\ 
\textbf{Q2:} Is United Kingdom an African country?\\ 
\textbf{A2:} No.\\ \\ 
\textbf{Merged-Q:} Did the first AI Safety Summit take place in an African country?\\ 
\textbf{Merged-A:} No\end{tabular} \\ 
\bottomrule
\end{tabular}
\end{adjustbox}
\caption{Data examples from CuQA dataset. For each example, one question is generated based on an unseen event and the other is generated based on model generated passage described in Section \ref{sec:data_collection}. The two questions and corresponding answers are then merged and post-processed to get the final question and answer.}
\label{tab:data_examples}
\end{table*}

\section{Experiment on FreshQA Dataset}

\subsection{Data Statistics}
% % Please add the following required packages to your document preamble:
% % \usepackage{multirow}
% \begin{table}[]
% \begin{tabular}{llc}
% \hline
% \multicolumn{3}{c}{count}                                                                     \\ \hline
% \multicolumn{1}{l|}{\multirow{2}{*}{\# hops}}        & \multicolumn{1}{l|}{multi-hop}   & 137 \\
% \multicolumn{1}{l|}{}                                & \multicolumn{1}{l|}{one-hop}     & 463 \\ \hline
% \multicolumn{1}{l|}{\multirow{4}{*}{effective year}} & \multicolumn{1}{l|}{before-2022} & 279 \\
% \multicolumn{1}{l|}{}                                & \multicolumn{1}{l|}{2022}        & 131 \\
% \multicolumn{1}{l|}{}                                & \multicolumn{1}{l|}{2023}        & 143 \\
% \multicolumn{1}{l|}{}                                & \multicolumn{1}{l|}{2024}        & 47  \\ \hline
% \multicolumn{1}{l|}{total}                           & \multicolumn{1}{l|}{-}            & 600 \\ \hline
% \end{tabular}
% \end{table}

\begin{table}[H]
\begin{adjustbox}{max width=0.48\textwidth}
\begin{tabular}{cc|cccc|c}
\toprule
\multicolumn{2}{c|}{\# hops} & \multicolumn{4}{c|}{effective year} & total \\ 
% \hline
multi-hop      & one-hop     & before-2022  & 2022  & 2023  & 2024 & -     \\ \midrule
137            & 463         & 279          & 131   & 143   & 47   & 600   \\ \bottomrule
\end{tabular}
\end{adjustbox}
\caption{Data statistics of FreshQA.}
\label{tab:append_freshqa_statistic}
\end{table}
We report the data statistics of the FreshQA dataset in Table \ref{tab:append_freshqa_statistic}.
Different fromthe  CuQA dataset that involves multi-hop reasoning for all instances, FreshQA is constructed to benchmark large language models' ability in addressing questions with time-changing knowledge. More than 77\% of questions are single-hop that requires no additional problem decomposition. The questions are split into four categories according to the effective year of the answers: \texttt{before-2022} ($46.50\%$), \texttt{2022} ($21.83\%$), \texttt{2023} ($23.83\%$), \texttt{2024} ($7.83\%$). 

\subsection{Analysis}

\begin{figure*}[!t]
    \centering
    \includegraphics[width=1.0\textwidth]{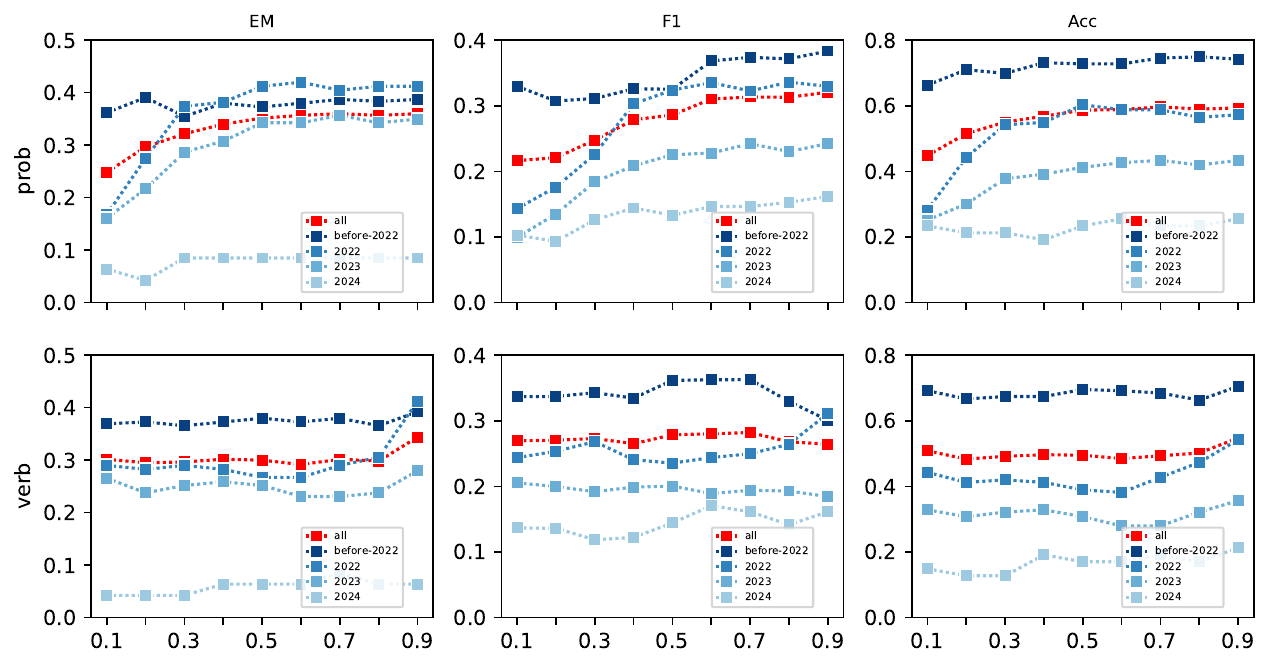}
    \caption{The performance of \texttt{1106} on FreshQA questions in different time-frames with varying $\alpha$ values. We fix $\beta$ as $0.1$ for the analysis. The first and second rows correspond to the performance with \textit{probability}- and \textit{verbalized}-based confidence scoring respectively.}
    \label{appendix-fig:time-frames-1106}
\end{figure*}

We present the performance details of \texttt{1106} on FreshQA by the time-frames of questions in Figure \ref{appendix-fig:time-frames-1106}.

\noindent \textbf{The performance of \textit{\texttt{Self-DC}} increases as the effective year of questions become earlier.} In general, the best performance is achieved on questions before 2022 and a decreasing trend is observed for more recent questions with both \textit{verb} and \textit{prob} confidence acquisition methods using \texttt{1106}. We also identified the same finding when using \texttt{4o-mini} with \textit{verb} method. This is not surprising as its training data ends up to September 2021\footnote{\url{https://platform.openai.com/docs/models/gpt-3-5-turbo}.}.

\section{More Analysis}

\subsection{Different Number of Retrieved Results}
\begin{figure}
    \centering
    \includegraphics[width=0.48\textwidth]{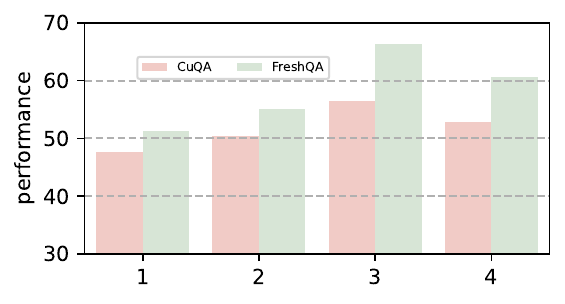}
    \caption{The performance of different number of retrieved results using $prob$ methods on \texttt{4o-mini}.}
    \label{fig:t_impacts_4o}
\end{figure}

We then set the number of retrieved results ranging from 1 to 4 to investigate the effects. Figure~\ref{fig:t_impacts_4o} shows the results. It is found that setting the number of retrieved results as 3 leads to the best performance for both of these two datasets, and the performance on FreshQA is more sensitive to the number of retrieved documents compared with CuQA.

% Despite setting the number of retrieved results as 3 leads to the best performance for both of these two datasets, there is a nuanced difference in the overall trend. For example, when the number is increased from 1 to 3, we can found it always leads to better performance with \texttt{1106}. However, the performance of \texttt{instruct} does not change when the number is less than 3. In addition, when the number exceeds 3, its performance decreases dramatically while the \texttt{1106} still achieves comparable performance compared with the number is 2. We attribute this to the more powerful capability of \texttt{1106} to capture the related information and filter noises.

\subsection{More Models.}

\begin{table}[h!]
\centering
\begin{tabular}{l|cccc}
\toprule
\textbf{Methods}        & \textbf{\#R} & \textbf{EM} & \textbf{F1} & \textbf{Acc} \\ 
\midrule
RR      & $n$          & 22.2        & 27.8        & 38.2         \\ 
ReFeed   & $2n$         & 24.8        & 17.2        & 33.8         \\ 
IRCoT  & $3n$         & 37.6        & 4.5     & 43.6         \\ 
\hline
\texttt{Self-DC} (Verb)  & $0$-$n$      & 23.6        & 26.9        & 35.8         \\ 
\texttt{Self-DC} (Prob)          & $0$-$n$      & 23.8        & 28.3        & 40.0         \\ 
\bottomrule
\end{tabular}
\caption{Performance results on Qwen2.5-7b-Instruct model.}
\label{tab:qwen}
\end{table}

We additionally run experiments on the Qwen2.5-7b-Instruct model by following the setting at main experiments. Table~\ref{tab:qwen} shows the final results. It is observed that our method still achieves better trade-off between effectiveness and efficiency.

\section{Demonstrations}
We mainly follow \citet{shao-etal-2023-enhancing} for prompt design. We list the used prompts and demonstrations for baselines in Tables \ref{tab:baseline-prompt-direct}-\ref{tab:baseline-prompt-selfask-continue} and the prompts for \textit{\texttt{Self-DC}} in Table \ref{tab:selfdc-prompt}. 
\begin{table}[H]
\small
    \centering
    \colorbox{purple!8}{
    \begin{tabular}{@{}p{7.2cm}}
    Please answer the following question with just a few words. \\\\Question: \{\texttt{question}\}\\
    The answer is
    \end{tabular}
    }
    \caption{Prompt for Direct Prompting baseline.}
    \label{tab:baseline-prompt-direct}
\end{table}

\begin{table}[H]
\small
    \centering
    \colorbox{purple!8}{
    \begin{tabular}{@{}p{7.2cm}}
    Please answer the following question with just a few words. Let's think step by step. \\\\
    Question: \{\texttt{question}\}\\
    The answer is
    \end{tabular}
    }
    \caption{Prompt for zero-shot Chain-of-thought baseline.}
    \label{tab:baseline-prompt-cot-zeroshot}
\end{table}

\begin{table}[H]
\small
    \centering
    \colorbox{purple!8}{
    \begin{tabular}{@{}p{7.2cm}}
    Refer to the passage below and answer the following question with just a few words. \\\\
    
    Passage: \{\texttt{passage}\}
    \\
    Question: \{\texttt{question}\}
    \\
    The answer is
    \end{tabular}
    }
    \caption{Prompt for Retrieve-then-read baseline. The \texttt{passage} comes from retrieval results.}
    \label{tab:baseline-prompt-rr}
\end{table}

% \begin{table}[H]
% \small
%     \centering
%     \colorbox{purple!8}{
%     \begin{tabular}{@{}p{7.2cm}}
%     Generate a background document from Wikipedia to answer the given question. \\
%     \\
%     \{\texttt{question}\}
%     \end{tabular}
%     }
%     \vspace{0.1cm}

%     \colorbox{purple!16}{
%     \begin{tabular}{@{}p{7.2cm}}
%     Refer to the passage below and answer the following question with just a few words. Passage: \{\texttt{passage}\} \\
%     \\
%     Question: \{\texttt{question}\}\\
%     \\
%     The answer is
%     \end{tabular}
%     }
%     \caption{Prompts for Generate-then-read baseline. Two steps are involved: \colorbox{purple!8}{background information generation} and \colorbox{purple!16}{document based question answering.}}
%     \label{tab:baseline-prompt-genread}
% \end{table}
\begin{table}[H]
\small
    \centering
    \colorbox{purple!8}{
    \begin{tabular}{@{}p{7.2cm}}
    Generate a background document from Wikipedia to answer the given question. \\
    \\
    \{\texttt{question}\}
    \end{tabular}
    }
    \vspace{0.1cm}

    \colorbox{purple!8}{
    \begin{tabular}{@{}p{7.2cm}}
    Refer to the passage below and answer the following question with just a few words. \\\\
    Passage: \{\texttt{passage}\} \\
    Question: \{\texttt{question}\}\\
    The answer is
    \end{tabular}
    }
    \caption{Prompts for Generate-then-read baseline.}
    \label{tab:baseline-prompt-genread}
\end{table}

\begin{table}[H]
\small
    \centering
    \colorbox{purple!8}{
    \begin{tabular}{@{}p{7.2cm}}
Quesion: Which country that has joined in 2023 Rugby World Cup in the final also held the 2023 FIFA Women's World Cup? \\
Are follow up questions needed here: Yes\\
Follow up: Which countries held the 2023 FIFA Women's World Cup?\\
Intermediate answer: Australia and New Zealand.\\
Follow up: Which countries have joined in 2023 Rugby World Cup in the final?\\
Intermediate answer: New Zealand, South Africa\\
So the final answer is: New Zealand\\
\\
Question: Is the country that held the 49th G7 summit same as where the 42nd G7 summit took place?\\
Are follow up questions needed here: Yes\\
Follow up: Where did the 42nd G7 summit take place?\\
Intermediate answer: Kruen, Bavaria, Germany\\
Follow up: Where did the 49th G7 summit take place?\\
Intermediate answer: Hiroshima, Japan\\
So the final answer is: No\\
\\
Question: What's the rate share of the 2023 elected president in the most developed South-East Asian countries?\\
Are follow up questions needed here: Yes\\
Follow up: Which country is the most developed in South-East Asia?\\
Intermediate answer: Singapore\\
Follow up: What's the rate share of the 2023 elected president in Singapore?\\
Intermediate answer: over 70 percent\\
So the final answer is: over 70 percent\\
% \\
%     Question: \{\texttt{question}\}\\
%     Are follow up questions needed here:
    \end{tabular}
    }
    \caption{Demonstrations for Self-Ask baseline on CuQA.}
    \label{tab:baseline-prompt-selfask-continue}
    \end{table}

\begin{table}[H]
\small
    \centering
    \colorbox{purple!8}{
    \begin{tabular}{@{}p{7.2cm}}
Quesion: Which country that has joined in 2023 Rugby World Cup in the final also held the 2023 FIFA Women's World Cup? \\
Let's think step by step\\
New Zealand and South Africa have joined in 2023 Rugby World Cup in the final, and Australia and New Zealand held the 2023 FIFA Women's World Cup.\\
So the answer is New Zealand\\
\\
Question: Is the country that held the 49th G7 summit same as where the 42nd G7 summit took place?\\
Let's think step by step\\
The 42nd G7 summit took place in Kruen, Bavaria, Germany and The 42nd G7 summit took place in Japan
So the answer is No\\
\\
Question: What's the rate share of the 2023 elected president in the most developed South-East Asian countries?\\
Let's think step by step\\
Singapore is the most developed and wealthy South-East Asia country, and the the rate share of the 2023 elected president in Singapore is over 70 percent.\\
So the answer is over 70 percent\\
% \\
% Question: \{\texttt{question}\}\\
% Let's think step by step
    \end{tabular}
    }
    \caption{Demonstrations for few-shot Chain-of-thought baseline on CuQA.}
    % \caption{Prompt for few-shot Chain-of-thought baseline}
    \label{tab:baseline-prompt-cot-fewshot}
\end{table}

\begin{table}[H]
\small
    \centering
    \colorbox{purple!8}{
    \begin{tabular}{@{}p{7.2cm}}
Passage: September 8 – October 28 – The 2023 Rugby World Cup is held in France, and New Zealand (the All Blacks) lost 11–12 to South Africa in the final at the Stade de France. 20 July – August 20 – The 2023 FIFA Women's World Cup is held in Australia and New Zealand. In the final, Spain wins 1–0 against England. \\
Quesion: Which country that has joined in 2023 Rugby World Cup in the final also held the 2023 FIFA Women's World Cup? \\
Let's think step by step\\
New Zealand and South Africa have joined in 2023 Rugby World Cup in the final, and Australia and New Zealand held the 2023 FIFA Women's World Cup.
So the answer is New Zealand\\
\\
Passage: The 42nd G7 summit took place in Kruen, Bavaria, Germany. The 49th G7 summit takes place in Hiroshima, Japan. Ukrainian president Volodymyr Zelenskyy arrives in Japan on the second day of the summit.\\
Question: Is the country that held the 49th G7 summit same as where the 42nd G7 summit took place?\\
Let's think step by step\\
The 42nd G7 summit took place in Kruen, Bavaria, Germany and The 42nd G7 summit took place in Japan
So the answer is No\\
\\
Passage: 1 September – 2023 Singaporean presidential election: Economist and former deputy prime minister Tharman Shanmugaratnam is elected president with a vote share of over 70 percent.\\
Question: What's the rate share of the 2023 elected president in the most developed South-East Asian countries?\\
Let's think step by step\\
Singapore is the most developed and wealthy South-East Asia country, and the the rate share of the 2023 elected president in Singapore is over 70 percent.\\
So the answer is over 70 percent
    \end{tabular}
    }
    \caption{Demonstrations for ITER-RETGEN baseline on CuQA.}
    \label{tab:baseline-prompt-retgen}
\end{table}

\begin{table}[H]
\small
    \centering
    \colorbox{black!8}{
        \begin{tabular}{@{}p{7.2cm}}
        Please read the question, give the answer and indicate your level of confidence. Use the following format to provide your answer and confidence level: \\ 
        \\
        Answer: [Your answer]\\
        Confidence (0-100): [Your confidence level, please only include the numerical number, e.g. 80]\% \\
        \\
        Note: The confidence level indicates the degree of certainty you have about your answer and is represented as a percentage. For instance, if your confidence level is 80\%, it means you are 80\% certain that your answer is correct and there is a 20\% chance that it may be incorrect. If you do not know the answer, simply output confidence as 0\%.\\
        \\
        Question: \{\texttt{question}\} Please answer this question and provide your confidence level. Note that the confidence level indicates the degree of certainty you have about your answer and is represented as a percentage. \\
        Answer:
        \end{tabular}
    }
    
     \vspace{0.1cm}
     
    \colorbox{green!8}{
    \begin{tabular}{@{}p{7.2cm}}
    Please read the question, divide the question into smaller, independent parts. By solving these individual sub-questions and combining their answers, you can derive the solution to the main question. Use the following format to provide your answer: \#1: [sub-question 1], \#2: [sub-question 2], ...\\
    \\
    Question: \{\texttt{question}\} \\
    Answer:
    \end{tabular}
    }
    \vspace{0.1cm}
    
    \colorbox{blue!8}{
    \begin{tabular}{@{}p{7.2cm}}
    Refer to the passage below and answer the following question with just a few words. Passage: \{\texttt{passage}\}\\
    \\
    Question: \{\texttt{question}\} \\
    The answer is
    \end{tabular}
    }

    \vspace{0.1cm}
    \colorbox{orange!8}{
    \begin{tabular}{@{}p{7.2cm}}
    Generate a background document from Wikipedia to answer the given question. \{\texttt{question}\} \\
    \\
    Refer to the passage below and answer the following question with just a few words. Passage: \{\texttt{passage}\}\\
    \\
    Question: \{\texttt{question}\} \\
    The answer is
    \end{tabular}
    }

    \vspace{0.1cm}
    \colorbox{red!10}{
    \begin{tabular}{@{}p{7.2cm}}
    Question: \{\texttt{question}\} \\
    \\
    Here are all related sub-questions and corresponding answers: \{\texttt{sub\_qas}\} \\
    \\
    According to answers of all related sub-quesions of the original question, please generate the final answer of the original question using a few words.
    \end{tabular}
    }
    \caption{Prompts for \textbf{\textit{Self-DC}}: \colorbox{black!8}{verbalize-based confidence acquisition}, \colorbox{green!8}{\textit{decompose}}, \colorbox{blue!8}{\textit{retrieve-then-read}}, \colorbox{orange!8}{\textit{generate-then-read}}, and \colorbox{red!10}{\textit{combine-sub-qas}}.}
    \label{tab:selfdc-prompt}
\end{table}

\end{document}